# Predicting Customer Churn: Extreme Gradient Boosting with Temporal Data

First-place Entry for Customer Churn Challenge in WSDM Cup 2018

Bryan Gregory
Seycor Consulting
bryan@seycor.com

## ABSTRACT

Accurately predicting customer churn using large scale time-series data is a common problem facing many business domains. The creation of model features across various time windows for training and testing can be particularly challenging due to temporal issues common to time-series data. In this paper, we will explore the application of extreme gradient boosting (XGBoost) on a customer dataset with a wide-variety of temporal features in order to create a highly-accurate customer churn model. In particular, we describe an effective method for handling temporally sensitive feature engineering. The proposed model was submitted in the WSDM Cup 2018 Churn Challenge and achieved first-place out of 575 teams.

## 1. INTRODUCTION

For many businesses, accurately predicting customer churn is critical to long-term success. Accurate prediction of churn probability drives many aspects of a business including proactive customer marketing, sales forecasting, and churn-sensitive pricing models. Therefore even slight improvements in accuracy can lead to dramatic improvements in profit.

Although customer churn models have existed in the business domain for decades, they have recently grown in complexity and accuracy as modern machine learning methods have advanced in recent years. Consequently, modern machine learning libraries such as extreme gradient boosting (XGBoost) can be applied to create very accurate models even on high dimensional data.

In addition, one of the most challenging aspects of creating churn models is handling issues associated with temporal data and ensuring all features are correctly accounting for time-shifts across the various time-windows that are used when training, cross-validating, and testing a machine learning model. In this model, we implemented a method for handling temporal feature engineering which was validated to be successful.

For the purposes of this paper, we participated in the WSDM 2018 Cup Customer Churn Challenge[1] to address the complicated task of accurately predicting customer churn using modern machine learning libraries such as XGBoost to explore methods of engineering temporally sensitive features as input for the churn model. The dataset for analysis was extracted from KKBOX, a leading music streaming service in Asia which utilizes a subscription based business model.

The goal for the challenge was to create the most accurate model possible for predicting customer churn using only the provided dataset. Model performance was evaluated using Log Loss. Computation time was not a considered factor, therefore any number of features could be created from the base data set, and any number and variety of learning algorithms could be used. The final model submissions were scored against a final test set and ranked according to log loss accuracy. The model described in this abstract scored a final log loss of .07974, achieving first-place out of 575 teams.

## 2. RELATED WORK

The success of the XGBoost library in creating highly-accurate models and thereby winning a large proportion of data science competitions has been well documented in recent years. Tree boosting has empirically proven to be a highly effective approach to predictive modeling, and it has shown success in across a wide array of problem domains. [1] Therefore we chose XGBoost as our primary learning algorithm when selecting which classifier to utilize for developing our customer churn model.

## 3. EXPERIMENTAL SETUP

### 3.1 Dataset

The dataset analyzed in this paper came from the WSDM Cup 2018 Challenge and was provided by KKBOX, a music streaming service. The dataset consisted of subscriber data from 3 distinct sources: user activity logs, transactions, and member data. 3 years of historical data were included.

User log data included a variety of information about subscriber activity by day, transaction data covered all payment and subscription information including renewals and cancellations, and member data contained demographic information about each subscriber such as birthdate and gender. All 3 data sources contained temporal elements, with user activity logs and transactions being a time-series and member data containing initial registration date of the subscriber as well as birthdate.

---

[1] https://www.kaggle.com/c/kkbox-churn-prediction-challenge

Target variable for our model was the "is_churn" field, which is a binary label generated via a provided Scala script. The criteria is defined as true if no renewal activity took place within the 30 days of a member's subscription expiration date (provided in transactions file).

### 3.2 Time Period Splits

For our experimentation, we split the dataset temporally into 3 separate time periods for training, cross-validation, and testing:
- Training: January (2017/01/01 – 2017/01/31)
- Cross Validation: February (2017/02/01 – 2017/2/28)
- Testing: March (2017/03/01 – 2017/03/31)

### 3.3 Evaluation Metrics

Model performance was evaluated using a standard log loss calculation on the target variable (churn probability).

$$logloss = -\frac{1}{N}\sum_{i=1}^{N}(y_i \log(p_i) + (1-y_i)\log(1-p_i))$$

Figure 1: Log Loss formula used for evaluating model performance

## 4. TEMPORAL FEATURE ENGINEERING

For our model input, we utilized two distinct methods of creating features from temporal elements. Each method was used for a distinct subset of features, with the goal of maximizing signal from each temporal feature while reducing false bias.

The majority of temporal features in our winning model utilized the relative refactoring method, with only a few using the absolute method, such as initial registration date and birth date. The absolute method is only used if it is thought that there is more signal in the absolute date (such as registering on January 2$^{nd}$, immediately after a holiday) than there is signal in the relative amount of time that has passed (29 days ago).

### 4.1 Relative Refactoring Method

The relative refactoring method is designed to create new features that are relative to the time period which is being modelled. In this way, temporal context is embedded in the feature for the model to learn. While relatively simple and intuitive, if not done carefully and consistently for all time periods modeled (training, cross-validation, and test data sets), then a feature that is known to contain important signal instead leads to a prediction bias which is simply noise.

The relative refactoring method involves mapping a date-driven feature into a new feature space which is not anchored in any particular time-series, but is instead relative to a selected point in time. This ensures a feature is relatively comparable across different time periods.

The formula for creating a relative refactored feature is simple:

a) Choose a unit of time measurement that is most appropriate for a given data element. For example, if a data element is a date, then choose days.

b) Choose a static point in time which represents the start of the time period being modelled, using the same unit of time measurement selected in part A. For example, if days were chosen, then select the first day of the time period being modelled

c) Create a new feature by taking the difference of the data element and the static point in time chosen in part B.

A simple example from the KKBOX dataset:

> The registration date for a sampled user is January 25th and the last login date for that user is January 30th. Using the relative refactoring method, we create new features from these two data elements which represent the number of days difference between the data elements and the start of a given prediction period. Therefore for the training dataset time period, we have a new "days since registration" feature which is calculated as 6 and a "days since login" feature which is calculated as 1. If we are predicting the cross-validation time period (Feb 1st – Feb 28th) and the user has not had any updated activity, then the "days since registration" feature is now calculated as 34 and the "days since last login" feature for that same user is now calculated as 29, even though the original date elements have not changed.

### 4.2 Absolute Method

This method simply involves converting the raw date field into an ordinal integer as using it as input for the model. For example, a registration date of 2017/01/01 is converted to 20170101. The ordinal integer conversion is necessary to for properly formatting the input to a numerical datatype for the XGBoost model. The conversion step was not necessary for the KKBOX dataset as the date fields given were already in integer format.

This method is effective when absolute date/time contains more signal than relative date/time. For example, in the KKBOX dataset, if there is a significant churn pattern in users who registered following a major holiday, absolute registration date is important. Applying the relative method to registration date would cause the model to incorrectly weight that churn behavior as a relative point in time for future months.

Both methods cannot be used together for the same time period on the training dataset. Using the absolute method on data with stronger relative signal causes the model to learn incorrectly, and vice-versa. One method or the other must be chosen for a given time period and feature. However, the two methods can be combined across different time period splits, based on which method is more effective for a given time period.

## 5. THE CHURN MODEL

### 5.1 Pre-Processing

Due to the size of the data being analyzed, the data files provided were loaded into a Microsoft SQL Server instance and cleaned and munged using T-SQL code stored as views. Cleaning steps included removing outliers, imputing nulls, and converting the integer date fields into date datatypes for ease of later feature engineering utilizing date differences (described in section 4.1).

Also, new churn labels were created for the training (January) and cross-validation (February) datasets using the Scala script provided by the competition sponsors. The prior churn labels provided by competition sponsors for those months were discarded.

### 5.2 Feature Engineering

The majority of feature engineering was performed in the T-SQL environment and imported into the Python code environment for input into XGBoost for model training.

Features were derived from data within the 3 data sources (user activity logs, transactions, and member data), as well as meta features (utilizing data derived from multiple sources) and higher order features (features that combined interaction between other features). An example of such a feature is the ratio of number of days since last login to payment plan days.

In total, 208 features were created and tested, with >80% of the features sourced from the user activity logs and the transactions datasets. Of the 208 total features, 76 were added to the final model.

### 5.3 Feature Selection

In total, 208 features were created and each was tested by cross-validation on the February dataset. A Python function was created to iteratively add each new feature to the existing validated features, re-train the XGBoost model, and record the new model accuracy resulting from the additional feature. Features which increased accuracy were added to the test model and others were discarded. Some features which exhibited an increase in model accuracy but were found to be overlapping based on very high correlation were manually discarded.

### 5.4 Classifier Training

The XGBoost library implemented in Python was selected as the primary classifier. Other learners tested included LightGBM, StackNet, and scikit-learn's implementation of Gradient Boosting Machines. Hyper-parameter optimization was performed based on cross-validation feedback for the month of February. To ensure fast cross-validation times even with high dimensionality as feature count grew, XGBoost's fast optimized histogram grower was used.

For the final model submission, a weighted average ensemble of XGBoost and LightGBM were used. Weights were assigned using competition leaderboard feedback and were 88% XGBoost base model and 12% LightGBM base model.

## 6. RESULTS AND FINDINGS

Feature importance, extracted from the final XGBoost model was found to have a surprisingly even spread amongst the many features, with no single feature dominating the model.

| Feature | Importance |
|---|---|
| payment_plan_days | 6.61% |
| ul_lastmo_last2wk_numunq_avg_diff | 4.56% |
| ul_all_numunq_sum_y | 4.19% |
| ul_mo1_mo2_trend | 3.99% |
| ul_lastprevmo_secs_sum | 3.98% |
| canc_per_payment_days | 3.81% |
| std_dev_numunq_prev_mo | 3.60% |
| ul_all_count_y | 2.95% |
| last_trx_gt1_no_cancel | 2.80% |
| last_ul_days_s | 2.63% |

Table 1: Top 10 feature importance, extracted from the final XGBoost model

Many of the most important features came from recent user log activity from prior months and were calculated by comparing various time periods to extract trends in user activity. For example, we created features which measured the change in user activity across many different time windows, such as the prior 2 weeks compared to the prior month and the prior month compared to the prior prior month. In addition, trend analysis was performed in many similar but distinct variations to ensure the model could learn as much as possible from the user activity domain. Various methods of measuring user activity were used such as sum, mean, max, etc. of a user's seconds played, number of unique songs played, and logins, aggregated over the various time windows.

Transaction features related to cancellations were found to be very relevant to customer churn as well. Examples of features engineered from this domain were: total number of cancellations in a user's transaction history, a boolean flag describing if prior month contained a cancellation transaction, and a customer's average cancellation rate per month[2].

---

[2]Some customers were found to have > 1 cancellation in their history as it is common for KKBOX customer to cancel one plan and later re-subscribe to a new plan.

Relative temporal features were also found to be important such as: days since last login, days since last significant usage, and days since last cancellation. The method used for engineering them are described in section 4.1.

| Rank | Team Name | Final Accuracy (Log Loss) |
|---|---|---|
| 1 | Bryan Gregory | 0.07974 |
| 2 | Swimming | 0.08926 |
| 3 | JonahWang | 0.09344 |
| 4 | 501 | 0.0966 |
| 5 | Alaric | 0.09664 |

Table 2: WSDM Cup 2018 Final Results

## 7. CONCLUSION AND FUTURE WORK

In summary, we used a supervised machine learning ensemble of decision trees, implemented in the modern XGBoost library, to build a highly accurate classification model for the purpose of predicting customer churn. Final accuracy was further boosted by using a weighted average ensemble of a base model trained using the LightGBM library and the primary XGBoost base model.

The overall accuracy achieved on the test dataset was a log loss of .07974, which very significantly out-performed known benchmarks, demonstrating that forecasting KKBOX subscriber churn with a significant level of accuracy is achievable using the methods described in this paper.

In addition, this model outscored all other models submitted for the challenge by other teams, and significantly outscored other XGBoost models submitted as well. This suggests that feature engineering was key in increasing the accuracy of the model, and suggests that our methods for crafting temporal features for the model were valid and successful.

Future work includes further parameter optimization of the XGBoost and LightGBM base models, stacking with StackNet models, and further exploration of additional feature engineering not yet tested.

## ACKNOWLEDGMENTS

We thank everyone associated with organizing and sponsoring the WSDM Cup 2018. Dataset was provided by KKBOX. Challenge was sponsored and managed by the 11th ACM International Conference on Web Search and Data Mining (WSDM 2018). Competition platform was hosted by Kaggle.